\documentclass[conference]{IEEEtran}

\IEEEoverridecommandlockouts

\def\BibTeX{{\rm B\kern-.05em{\sc i\kern-.025em b}\kern-.08em
    T\kern-.1667em\lower.7ex\hbox{E}\kern-.125emX}}
\pdfoutput=1 

\usepackage[table,usenames,dvipsnames]{xcolor}
\usepackage{annotations}
\usepackage{enumitem}

\usepackage{pifont}
\definecolor{cadmiumgreen}{rgb}{0.0, 0.42, 0.24}
\usepackage{multirow}
\usepackage{setspace} 
\usepackage{soul}
\usepackage{mathtools}
\usepackage{amsmath,amssymb,amsthm}

\usepackage{setspace}
\usepackage{nicefrac}
\usepackage{siunitx}
\usepackage{array,framed}
\usepackage{booktabs}
\usepackage{
  color,
  float,
  epsfig,
  wrapfig,
  graphics,
  graphicx,
}
\usepackage{setspace}
\usepackage{amsfonts}
\usepackage{latexsym,fancyhdr,url}
\usepackage[linesnumbered,ruled,vlined]{algorithm2e}
\usepackage{algpseudocode}
\usepackage{graphics}
\usepackage{xparse} 
\usepackage{xspace}
\usepackage{multirow}
\usepackage{csvsimple}
\usepackage{balance}
\usepackage{listings}
\usepackage{mathtools}

\usepackage{caption}
\DeclareMathAlphabet{\mathcal}{OMS}{cmsy}{m}{n}

\usepackage{multirow}
\usepackage{booktabs}
\usepackage{tabularx}
\newcolumntype{L}[1]{>{\raggedright\let\newline\\\arraybackslash\hspace{0pt}}m{#1}}
\newcolumntype{C}[1]{>{\centering\let\newline\\\arraybackslash\hspace{0pt}}m{#1}}
\newcolumntype{R}[1]{>{\raggedleft\let\newline\\\arraybackslash\hspace{0pt}}m{#1}}

\usepackage{etoolbox}
\patchcmd{\quote}{\rightmargin}{\leftmargin 2em \rightmargin}{}{}

\usepackage{listings}

\newcommand{\listingcaption}[1]%
{%
\refstepcounter{lstlisting}{\hspace{-15pt} Listing \thelstlisting: #1}
}%
\newlength{\MaxSizeOfLineNumbers}%
\usepackage[ruled]{algorithm2e}

\SetAlCapNameFnt{\footnotesize}
\SetAlCapFnt{\footnotesize}
\SetAlCapHSkip{0pt}
\IncMargin{-\parindent}
\usepackage{algorithmicx}
\usepackage[export]{adjustbox}
\usepackage[caption=false,font=footnotesize,labelformat=simple]{subfig} 

\usepackage{enumitem}
\newcommand{\rt}[1]{\color{red}{#1}}
\newcommand{\gt}[1]{{\color{NavyBlue}{#1}}}
\newcommand{\vt}[1]{{\color{Fuchsia}{#1}}}
\newcommand{\bt}[1]{{\color{Sepia}{#1}}}
\newcommand{\mat}[1]{{\color{PineGreen}{#1}}}
\newcommand{\bst}[1]{{\color{Bittersweet}{#1}}}
\newcommand{\fgt}[1]{{\color{WildStrawberry}{#1}}}
\usepackage{acronym}
\acrodef{IC}{integrated circuit}
\acrodef{EDA}{electronic design automation}
\acrodef{HDL}{hardware description language}
\acrodef{AIG}{and-inverter-graph}
\acrodef{ML}{machine learning}
\acrodef{IP}{intellectual property}
\acrodef{RTL}{register transfer level}
\acrodef{DAG}{directed acyclic graph}
\acrodef{GCN}{graph convolutional network}
\acrodef{SoC}{system-on-chip}
\acrodef{DNN}{deep neural network}
\acrodef{MDP}{Markov decision process}
\acrodef{CNN}{convolutional neural network}
\acrodef{MSE}{mean-square error}
\acrodef{SA}{simulated annealing}
\acrodef{SOTA}{state-of-the-art}
\acrodef{GNN}{graph neural network}
\acrodef{KPA}{key prediction accuracy}
\acrodef{OMLA}{oracle-less machine learning attack}
\acrodef{PPA}{power, performance, and area}
\acrodef{Acc}{accuracy}
\acrodef{wolog}{without loss of generality}

\usepackage{booktabs}    
\usepackage{tabularx}
\usepackage{booktabs}    
\usepackage{multirow}
\usepackage{graphicx}
\usepackage{xparse}
\newcommand{\bnm}{\begin{newmath}}
\newcommand{\enm}{\end{newmath}}

\newcommand{\bea}{\begin{eqnarray*}}%
\newcommand{\eea}{\end{eqnarray*}}%

\newcommand{\bne}{\begin{newequation}}
\newcommand{\ene}{\end{newequation}}

\newcommand{\bal}{\begin{newalign}}
\newcommand{\eal}{\end{newalign}}

\newenvironment{newalign}{\begin{align}%
\setlength{\abovedisplayskip}{4pt}%
\setlength{\belowdisplayskip}{4pt}%
\setlength{\abovedisplayshortskip}{6pt}%
\setlength{\belowdisplayshortskip}{6pt} }{\end{align}}

\newenvironment{newmath}{\begin{displaymath}%
\setlength{\abovedisplayskip}{4pt}%
\setlength{\belowdisplayskip}{4pt}%
\setlength{\abovedisplayshortskip}{6pt}%
\setlength{\belowdisplayshortskip}{6pt} }{\end{displaymath}}

\newenvironment{newequation}{\begin{equation}%
\setlength{\abovedisplayskip}{4pt}%
\setlength{\belowdisplayskip}{4pt}%
\setlength{\abovedisplayshortskip}{6pt}%
\setlength{\belowdisplayshortskip}{6pt} }{\end{equation}}

\newcounter{ctr}

\newcounter{mytable}
\def\mytable{\begin{centering}\refstepcounter{mytable}}
\def\endmytable{\end{centering}}

\newcounter{myfig}
\def\myfig{\begin{centering}\refstepcounter{myfig}}
\def\endmyfig{\end{centering}}

\newlength{\saveparindent}
\newlength{\saveparskip}
\newcommand{\E}{{\rm I\kern-.3em E}}

\renewcommand{\eqref}[1]{\mbox{Equation~(\ref{#1})}}

\def \part {part}

\renewcommand{\paragraph}[1]{\vspace*{6pt}\noindent\textbf{#1}\;}

\def \blackslug{\hbox{\hskip 1pt \vrule width 4pt height 8pt
    depth 1.5pt \hskip 1pt}}
\def \qed{\quad\blackslug\lower 8.5pt\null\par}

\newcounter{mynote}[section]

\newcommand\ignore[1]{}

\newcounter{rcnote}[section]

\newcounter{mrnote}[section]

\newcounter{fknote}[section]

\newcounter{anote}[section]

\DeclareMathSymbol{\mlq}{\mathord}{operators}{``}
\DeclareMathSymbol{\mrq}{\mathord}{operators}{`'}

\newcommand{\rhf}[2]{R_{f, \gamma}}

\DeclareDocumentCommand{\edist}{o o}{
  \ensuremath{
    \IfNoValueTF{#1}{{d}}{{\sf d}(#1,#2)}
  }
}

\newcommand{\olrk}[1]{\ifx\nursymbol#1\else\!\!\mskip4.5mu plus 0.5mu\left(\mskip0.5mu plus0.5mu #1\mskip1.5mu plus0.5mu \right)\fi}

\NewDocumentCommand{\indseq}{ O{1} O{r} }{{#1}\ldots {#2}}

\definecolor{mygreen}{rgb}{0,0.6,0}
\definecolor{mygray}{rgb}{0.5,0.5,0.5}
\definecolor{mymauve}{rgb}{0.58,0,0.82}

\makeatletter
\let\old@lstKV@SwitchCases\lstKV@SwitchCases
\def\lstKV@SwitchCases#1#2#3{}
\makeatother
\usepackage{lstlinebgrd}
\makeatletter
\let\lstKV@SwitchCases\old@lstKV@SwitchCases

\lst@Key{numbers}{none}{%
    \def\lst@PlaceNumber{\lst@linebgrd}%
    \lstKV@SwitchCases{#1}%
    {none:\\%
     left:\def\lst@PlaceNumber{\llap{\normalfont
                \lst@numberstyle{\thelstnumber}\kern\lst@numbersep}\lst@linebgrd}\\%
     right:\def\lst@PlaceNumber{\rlap{\normalfont
                \kern\linewidth \kern\lst@numbersep
                \lst@numberstyle{\thelstnumber}}\lst@linebgrd}%
    }{\PackageError{Listings}{Numbers #1 unknown}\@ehc}}
\makeatother

\usepackage{lstlinebgrd} 

\lstdefinestyle{mystyle}{
    language=verilog,
    frame=single,
    basicstyle=\scriptsize\ttfamily,
    aboveskip=1em, 
    belowskip=1em, 
    captionpos=b,  
}

\DeclareCaptionLabelFormat{mystyle}{Prompt #2}
\usepackage{comment}
\usepackage{listings}

\usepackage{circledsteps}

\usetikzlibrary{patterns}

\begin{document}
%
\title{Towards the Imagenets of ML4EDA}


\author{\IEEEauthorblockN{Animesh B. Chowdhury\IEEEauthorrefmark{1},
Shailja Thakur\IEEEauthorrefmark{1},
Hammond Pearce\IEEEauthorrefmark{2}, 
Ramesh Karri\IEEEauthorrefmark{1} and
Siddharth Garg\IEEEauthorrefmark{1}}
\IEEEauthorblockA{\IEEEauthorrefmark{1}New York University, USA}
\IEEEauthorblockA{\IEEEauthorrefmark{2}University of New South Wales, Australia}}


\maketitle

\begin{abstract}
Despite the growing interest in ML-guided EDA tools from RTL to GDSII, there are no standard datasets or prototypical learning tasks defined for the EDA problem domain. Experience from the computer vision community suggests that such datasets are crucial to spur further progress in ML for EDA. Here we describe our experience curating two large-scale, high-quality datasets for Verilog code generation and logic synthesis. The first, VeriGen, is a dataset of Verilog code collected from GitHub and Verilog textbooks. The second, OpenABC-D, is a large-scale, labeled dataset designed to aid ML for logic synthesis tasks. The dataset consists of 870,000 And-Inverter-Graphs (AIGs) produced from 1500 synthesis runs on a large number of open-source hardware projects. In this paper we will discuss challenges in curating, maintaining
and growing the size and scale of these datasets. We will also touch upon questions of dataset quality and security, and
the use of novel data augmentation tools that are tailored for the hardware domain.
\end{abstract}


%
\IEEEpeerreviewmaketitle

\section{Introduction}

Digital design, especially from RTL to GDSII, has seen a surging interest in the application of Machine Learning (ML) to enhance the Electronic Design Automation (EDA) process. Yet, as the field progresses, there is a noticeable gap: the lack of standard datasets or typical learning tasks for the EDA domain. In areas like computer vision, standard datasets~\cite{imagenet} have played a pivotal role in advancing research. Drawing inspiration from such success, there is a clear need for similar resources in ML for EDA to drive further innovation.

We embarked on the journey to curate two groundbreaking datasets aimed at Verilog code generation and logic synthesis. Firstly, VeriGen emerges from an extensive collection of Verilog code sourced from GitHub repositories and textbooks. It acts as a reservoir of data for training models to generate accurate and efficient Verilog code. Secondly, OpenABC-D stands out as a significant contribution for logic synthesis. Consisting of a whopping 870,000 And-Inverter-Graphs (AIGs) derived from 1500 synthesis runs, this dataset encompasses a vast array of open-source hardware projects.

This paper discusses  the nuances of assembling these datasets,  challenges in their curation, maintenance, and expansion. Besides the sheer scale of these datasets, we will  explore  concerns like data quality, security, and the adoption of innovative data augmentation tools designed for the hardware domain. With the unveiling of OpenABC-D and VeriGen datasets, we seek to fill the void of datasets for ML for EDA, empowering researchers  to innovate in design, synthesis, and optimization, driving the next wave of digital  design.

\section{Dataset for Verilog Code Generation}

\noindent\textbf{Trends in Hardware Design}: Typically, designers rely on Hardware Description Languages (HDLs) like Verilog and VHDL to describe hardware architectures and their operational behaviors. These HDLs are the foundation of digital designs. But, their translation from a generic specifications in natural languages, such as English-language requirements documents, is a perennial challenge. Converting  natural language specifications to HDL is done by hardware engineers, a process that is time-consuming, intricate and susceptible to errors. 

There have been concerted efforts to refine the HDL design process and elevate its efficiency. One notable strategy is high-level synthesis, which offers developers the luxury of detailing functionality using more familiar languages such as C. However, while this method does reduce the immediate complexities of the design process, it compromises hardware efficiency. Some researchers have explored modernizing HDLs through the integration of features and constructs common in software development. A prime example is Chisel, which is rooted in Scala, as introduced by Bachrach et al.~\cite{bachrach_chisel_2012}.  

\noindent\textbf{Automating HDL Design through Large Language Models (LLMs)}: With the challenges associated with conventional methods, the exploration of Artificial Intelligence (AI) or ML-based tools as an alternative for translating specifications to HDL has garnered attention. Recently, there hass been an intriguing evolution in the form of Large Language Models (LLMs)\cite{chen_evaluating_2021}, popularized by commercial offerings like GitHub Copilot and OpenAI ChatGPT~\cite{chatgpt}, appear as a promising candidate for this machine translation application. 


Built on the transformer architecture~\cite{vaswani_attention_2017}, LLMs can parse  natural and structured languages. LLMs predict next tokens based on a series of input tokens producing detailed text from a given prompt. Thus, when fed a technical specification or a coding directive, LLMs can generate the corresponding code. 
The potential of LLMs in converting natural language directives into HDL code is the rationale for our study. However, benchmarks mostly focus on software languages like Python.

\noindent\textbf{Dearth of Open Datasets is the Major Challenge:} An impediment in the adoption of LLMs for HDL generation is the scarcity of comprehensive and open datasets~\cite{li2023starcoder,pearce_asleep_2021}. For effective model training and evaluation for niche areas like Verilog code generation, extensive, and diverse datasets are necessary. Furtherm there is no open dataset tailored for training and evaluating LLMs for Verilog. Commercial LLMs like GitHub Copilot are susceptibility to errors when generating Verilog code due to the absence of rich and varied training data~\cite{pearce_asleep_2021}. DAVE ~\cite{pearce_dave_2020}, resorted to synthetic Verilog datasets. However,  models trained on limited and synthetic data falter when presented with unfamiliar tasks. Such challenges underline the need for real-world Verilog datasets to evaluate the efficacy of LLMs in Verilog code generation. Our research makes the following contributions:
\begin{enumerate}
    \item \textbf{VeriGen Dataset:} We amassed the most extensive Verilog training corpus for LLM training.
    \begin{enumerate}
        \item \textit{Code Corpus:} We have combined available open-source Verilog code, creating an extensive Verilog training corpus for LLM training. This corpus has high quality, diverse and real-world challenges.
        \item \textit{Textbook Corpus:}  Verilog code corpus was augmented with a curated selection from Verilog textbooks. 
    \end{enumerate}
    \item \textbf{VeriGen Evaluative Framework:} We craft a set of Verilog coding challenges with varied difficulty and devise test benches and benchmark syntactic accuracy and functional correctness of the Verilog generated by LLMs.
    \item \textbf{Open Research Advocacy:} We are making our extensive training and evaluation resources, available to the global community, link (\textbf{\url{https://github.com/shailja-thakur/VGen}} and \textbf{\url{https://github.com/NYU-MLDA/OpenABC}}). This will not only empower current researchers but will catalyze further innovations.
\end{enumerate}

\subsection{VeriGen Dataset}
\noindent\textbf{Verilog Code Corpus:} The primary corpus for training was derived from open-source Verilog code found in public GitHub repositories. Using Google BigQuery, a snapshot of more than 2.8 million Verilog repositories from GitHub was acquired. The collection process involved searching for specific keywords like "Verilog" and targeting files with the ``\texttt{.v}'' extension. Following the initial retrieval, there was a rigorous filtering process. Files were de-duplicated based on MinHash and Jaccard similarity metrics, and only ``\texttt{.v}'' files containing both ``\texttt{module}'' and ``\texttt{endmodule}'' statements were retained. This meticulous process ensured the exclusion of any irrelevant ``\texttt{.v}'' file without any Verilog content. The ultimate count came to around 65k ``\texttt{.v}'' files, aggregating to approximately 1GB. Large files, particularly those with a character count exceeding 20K, were further filtered to ensure the quality of the dataset, leading to a final training corpus from GitHub that comprised around 50K files or roughly 300MB.

\noindent\textbf{Verilog Books Corpus:} To augment the primary training dataset, 70 Verilog-centric textbooks were procured in PDF format from an online e-library. The text extraction process was executed using the Python-based tool "pymuPDF" , which employs optical character recognition. However, the text quality was contingent upon the PDF's quality, leading to challenges like the misformatting of lines, unrecognized logical operators, and misplaced headers/footers.


\begin{lstlisting}[language=TeX, basicstyle=\scriptsize]
module(.*\n*\s*\t*)(\()((?!module)(?!endmodule).*\W*)*endmodule
\end{lstlisting}



\begin{figure}[t]
\setlength\abovecaptionskip{-0.05\baselineskip}
    \centering
    \includegraphics[width=0.6\linewidth]{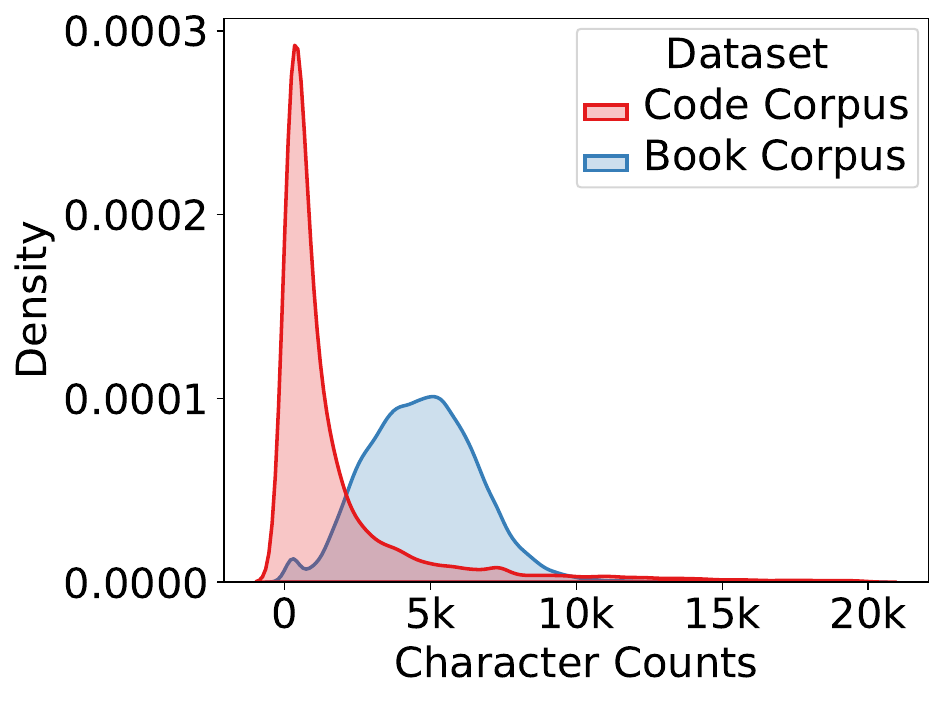}
    \vspace{-0.2em}
    \caption{Character count distribution of Code and text corpus}
    \vspace{-0.3em}
    \label{fig:dataset}
\end{figure}

Figure~\ref{fig:dataset} shows the distribution of character across book and code corpus. The figure reveals that the code corpus exhibits greater variability in length and potentially shorter samples, indicating that code snippets are often more concise due to the constraints of programming constructs. The broad range of character counts in both corpora emphasizes the diversity of content encompassed by the dataset.

\subsection{Evaluation Framework}

To comprehensively gauge the capabilities of Large Language Models (LLMs) in producing high-quality Verilog code, an elaborate evaluation framework has been designed. While this framework primarily targets models trained on the VeriGen Dataset, its versatile design ensures compatibility with any LLM adept at Verilog code generation. The foundation of this framework is built upon a two-fold strategy:
\begin{enumerate}
    \item A meticulously curated problem set, referred to as Set I, supported by hand-designed test benches.
    \item A rigorous end-to-end pipeline that evaluates the Verilog code for functional correctness.
\end{enumerate}

\noindent\textbf{Problem Set:} This encompasses 17 singular Verilog challenges (refer Table~\ref{tbl:problem_set}), inspired by classroom exercises and examples from the HDLBits platform. Each problem is categorized by a specific difficulty level, \textit{Basic Level:} tasks probe into foundational concepts, with challenges like designing a simple wire (Problem 1) or a 2-input AND gate (Problem 2), \textit{Intermediate Level:} challenges delve deeper, they encompass designs such as a half adder (Problem 5), a 1-to-12 counter (Problem 6), and an LFSR with specific taps (Problem 7), and \textit{Advanced Level:} tests the depth of expertise with tasks like crafting an FSM to recognize the sequence 101 (Problem 15) or the intricate ABRO FSM from Potop-Butucaru, Edwards, and Berry's work (Problem 17). The variation in challenges ensures an exhaustive test of the LLM's understanding and capability in diverse design concepts.

An illustrative example is depicted in Prompt~\ref{mylisting}, applied to the FSM problem. This prompt features a detailed level of description. The line 1 contain the initial comments, which describe the problem function. This section uses signal names and high-level specifications. The following segment (lines 2-4) includes the module header skeleton, which comprises the module's name and input/output data. And lines5-15 holds additional line of comments describing the logic in detail.

\begin{table}[t]
\caption{Problem set}
\centering
\footnotesize
\begin{tabular}{cll} 
\toprule
\textbf{Prob. \#} & \textbf{Difficulty} & \textbf{Description} \\ 
\midrule
1 & Basic & A simple wire \\ 
2 & Basic & A 2-input and gate \\
3 & Basic &  A 3-bit priority encoder \\
4 & Basic & A 2-input multiplexer \\
5 & Intermediate & A half adder  \\
6 & Intermediate & A 1-to-12 counter  \\
7 & Intermediate & LFSR with taps at 3 and 5 \\
8 & Intermediate & FSM with two states \\
9 & Intermediate & Shift left and rotate \\
10 & Intermediate & Random Access Memory\\
11 & Intermediate & Permutation  \\
12 & Intermediate & Truth table \\
13 & Advanced & Signed 8-bit adder with overflow \\
14 & Advanced & Counter with enable signal \\
15 & Advanced & FSM to recognize `101' \\
16 & Advanced & 64-bit arithmetic shift register \\
17 & Advanced & ABRO FSM$^{*}$ \\
\bottomrule
\end{tabular}
\footnotesize{$^{*}$from Potop-Butucaru, Edwards, and Berry's ``Compiling Esterel''}
\label{tbl:problem_set}
\end{table}

\noindent\textbf{LLM Inference Mechanism:} The process starts with an input prompt from Problem Set. The completed code, upon reaching the keywords ``\texttt{end}'' or ``\texttt{endmodule}'', is relayed to the evaluation harness to assess both its compilation and functional integrity. Key Parameters for Inference includes,
\noindent\textit{}{Prompts:} Three levels of prompts have been formulated: low (L), medium (M), and high (H). The depth and detail of information increase from L to H, with H resembling pseudo-code instructions. For instance, in Prompt\ref{mylisting}, the prompt for Problem 15 demonstrates the hierarchy—L comprises lines 1-4, M includes up to line 8, and H spans until line 15.

\noindent\textit{Sampling Temperature $(t)$}  influences the LLM's creativity. A higher value means riskier and more diverse completions. We've experimented with values like $t \in {0.1,0.3.0.5,0.7,1}$.
\noindent\textit{Completions per Prompt $(n)$} The LLM produces $n$ completions for every prompt. The values of 
$n$ we have considered include {1,10,25}, although with certain models, some of these options are skipped due to constraints.
\noindent\textit{Max Tokens:} An integral factor, the max\_tokens parameter caps the number of tokens generated in a single completion. Different models have different max\_token limits.

\definecolor{blond}{rgb}{0.98, 0.94, 0.75}
\definecolor{lightergray}{RGB}{220, 220, 220}

\begin{lstlisting}[language=verilog,basicstyle=\scriptsize\ttfamily,frame=single,
% linebackgroundcolor={\ifnum\value{lstnumber}>4
%                 \ifnum\value{lstnumber}<9
%                     \color{blond}
%                 \fi
%             \fi 
%             \ifnum\value{lstnumber}>8
%                 \ifnum\value{lstnumber}<16
%                     \color{lightgray}
%                 \fi
%             \fi},
            caption=Example - Problem 15: FSM to recognize `101',
    label=mylisting]
// This is a finite state machine that recognizes the sequence 101 on the input signal x. 
module adv_fsm(input clk, input reset, input x, output z); 
reg [1:0] present_state, next_state;
parameter IDLE=0, S1=1, S10=2, S101=3;
// output signal z is asserted to 1 when present_state is 
// S101
// present_state is reset to IDLE when reset is high, 
// otherwise it is assigned next state
// if present_state is IDLE, next_state is assigned S1 if 
// x is 1, otherwise next_state stays at IDLE
// if present_state is S1, next_state is assigned S10 if
// x is 0, otherwise next_state stays at IDLE 
// if present_state is S10, next_state is assigned S101 if 
// x is 1, otherwise next_state stays at IDLE 
// if present_state is S101, next_state is assigned IDLE

\end{lstlisting}

\noindent\textbf{Test Benches:}
An intrinsic part of Problem Set, the test benches have been purposefully created to validate the functional accuracy of each Verilog solution. They have been designed to be exhaustive, especially for basic challenges, but are more akin to unit tests for some of the complex ones. This approach ensures a balanced evaluation that is both rigorous and time-efficient.
Utilizing Icarus Verilog v11.0, the Verilog codes are compiled and simulated. 

\begin{figure}[t]
\setlength\abovecaptionskip{-0.05\baselineskip}
    \centering
    \includegraphics[width=0.9\linewidth]{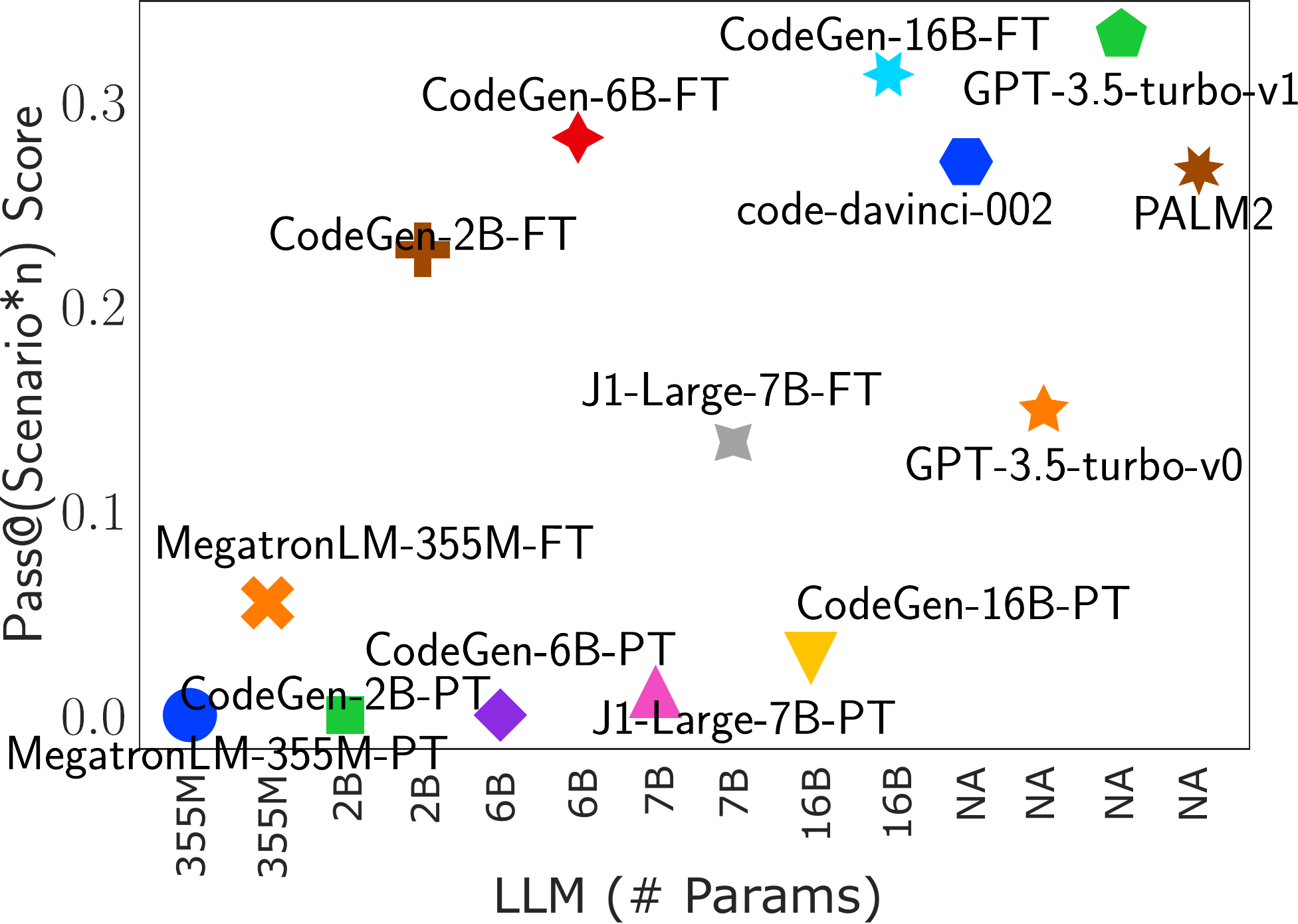}
    \vspace{0.3em}
    \caption{Performance Comparison: Pass@(Scenario*$n$) vs. Inference Time; Pass@(scenario*$n$) for scenarios passing test benches across problem difficulties and description levels. Higher is better; Inference Time is the time time from submitting a query to getting a response from LLM.}
    \vspace{-0.3em}
    \label{fig:description-difficulty}
\end{figure}

    

\subsection{Benchmarking the performance of LLMs for Verilog}

We assess the quality of generated code using the problem sets detailed in~\ref{tbl:problem_set}. A "scenario" denotes an amalgamation of problems across various difficulty levels and description tiers. 

We characterize the model performance with the Pass@$k$ metric, where $k$ is the number of problems in a \textit{scenario} times $n$, the number of suggestions by LLM per problem. A higher Pass@$k$ indicates a relatively `superior' result. In our analysis, we present Pass@$k$ for a set of $n=10$ completions. For functional tests, this metric represents the ratio of the $k$ code samples that pass these tests. Additionally, we report the inference time for each query made to the Language Models, which includes communication time with remote servers when applicable. 
Figure~\ref{fig:description-difficulty} shows the performance across a range of LLMs including both fine-tuned (FT) and pre-trained (PT) variants: Megatron-LM~\cite{shoeybi_megatron-lm_2020}, Jurassic (J1-Large)\cite{ai21_jurassic-1_2021}, CodeGen\cite{nijkamp_conversational_2022}, ChatGPT (GPT-3.5-turbo)~\cite{ye2023comprehensive}, and Codex (code-davinci-002)~\cite{chen_evaluating_2021}, as well as models with varying size denoted by no of parameters in Figure~\ref{fig:description-difficulty}. It is important to note that the reported results stem from fine-tuned models trained solely on the GitHub corpus. Its notable that model with higher number of parameters generally exhibits higher performance. While, larger models like GPT-3.5-turbo with special instructions (denoted as GPT-3.5-turbo-v1)  tend to exhibit higher scores, they demand more computational resources, as these models often require more time for inference.
Interestingly, the performance of the fine-tuned CodeGen-16B-FT stands out as exceptional, surpassing all other LLMs across the various problem scenarios. In general, the fine-tuned LLMs consistently outperform their pre-trained counterparts. And a case study, further fine-tuning with book corpus shows an improvement in the overall performance. codegen-2B-FT upon fine-tuning with book corpus shows a 3.5\% increase in the functionally correct  code. 

\section{Dataset for PPA prediction}
\label{sec:openabcd}

Power, performance, and area (PPA) represent critical metrics for appraising RTL design quality. The core of this evaluation lies in logic synthesis, a transformative process that converts RTL design into an equivalent netlist constructed from Boolean logic gates. Throughout the Electronic Design Automation (EDA) pipeline, the aim is to maximize PPA metrics, driving them towards optimal levels. The particular significance of logic synthesis stems from its role as the inaugural phase within the sequence of EDA stages, ultimately culminating in the final GDSII layout. The synthesized output bears substantial consequences, exerting a direct impact on IC dimensions, power efficiency, and computational speed.


\noindent\textbf{Logic synthesis} is $\Sigma_{p}^2$-Hard problem~\cite{umans1999hardness,buchfuhrer2011complexity} often tackled using heuristic approaches developed jointly by academia and industry~\cite{amaru2017logic}. It transforms a design's functional representation expressed as and-inverter-graphs (AIGs) through a sequence of logic minimization heuristics. These heuristics range from removing redundant nodes to refining Boolean formulas~\cite{abc}. The sequence of heuristics (``recipe") applied significantly impacts the results. Skilled engineers manually refine such recipes through iterative trial and error to achieve the best PPA outcomes. The formal problem is: \textit{Find the sequence of sub-graph optimization steps for an optimal AIG graph representing a hardware design's boolean functionalities.}

\noindent\textbf{ML for logic synthesis}: ML has rejuvenated interest in logic synthesis.
The key question revolves around leveraging past data for decisions in future instances. ML approaches predict:
\begin{enumerate}
\item recipe quality: \cite{cunxi_dac} employs a classifier to distinguish ``angel" recipes from ``devil" recipes for IPs. 
\item  optimal optimizer:  \cite{lsoracle}  is a model to determine the AIG/MIG representation for sub-circuit optimization.
\item synthesis recipe via reinforcement learning (RL): Recent works~\cite{firstWorkDL_synth,drills,mlcad_abc,cunxi_iccad} cast the challenge as a Markov Decision Process (MDP), where future AIG depends on the present AIG state and synthesis transformation (action). However, for cross-IP efficacy, the agent must explore diverse state-action pairs before transitioning to ``exploit" phase, dealing with unseen data.
\end{enumerate} 

Despite promise, comparing diverse techniques encounters challenges due to lacking standardized datasets, uniform configurations, and common benchmarks. Evaluations span varied datasets, hampering comprehensive approach assessment. RL methods exhibit high sample complexity, impeding extensive pre-training of RL agents due to synthesis-associated costs.

\textbf{OpenABC-D dataset}: We next introduce OpenABC-D~\cite{openabc-d,chowdhury2022bulls},  an extensive labeled dataset created by synthesizing open-source designs with ABC, a leading open-source logic synthesis tool~\cite{abc}. This dataset offers a valuable resource for developing, evaluating, and benchmarking machine learning-aided logic synthesis techniques. OpenABC-D includes 870,000 AIGs from 1500 synthesis runs, with labels like optimized node counts and delay metrics. Area and delay data are obtained by mapping AIGs using the nangate45nm library. We also offer an open-source framework for generating labeled data via synthesis runs on IPs using different synthesis recipes.


\begin{figure}[b!]
\centering
\includegraphics[width=0.8\columnwidth]{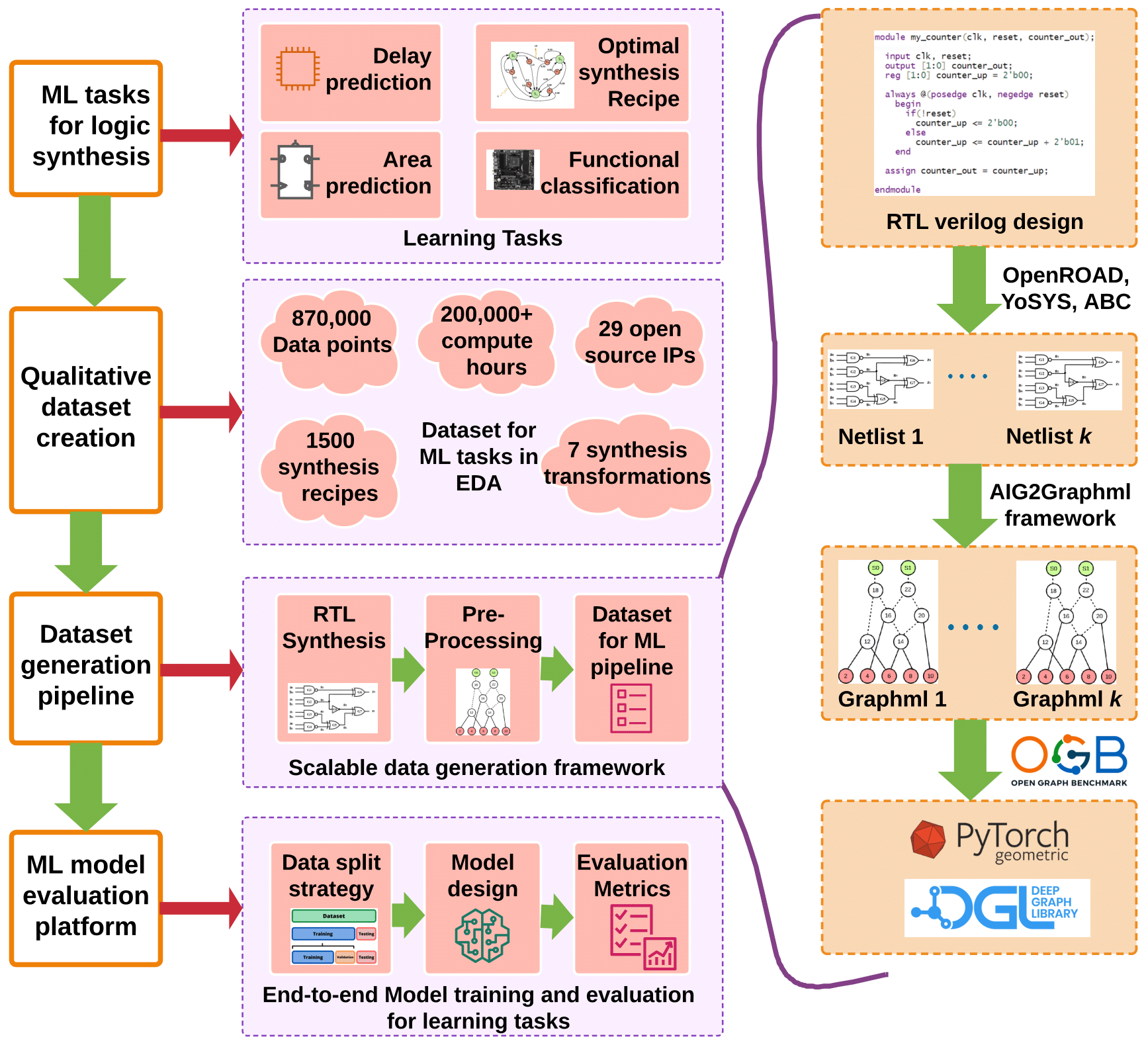}
\caption{OpenABC-D framework}
\label{fig:openabcdframework}
\end{figure}

\subsection{Data Synthesis Workflow}
\label{subsec:openabcd-dsw}

OpenABC-D framework uses open source EDA tools for  data generation, but this demands substantial computational resources for synthesis, pre-processing, and ML-ready format conversion. OpenABC-D is applied to open source IPs, sharing the dataset and highlighting contributions and development challenges. It is an end-to-end framework that boosts ML research in synthesis (see Fig.~\ref{fig:openabcdframework}).

We use OpenROAD v1.0~\cite{ajayi2019openroad} EDA for logic synthesis, with Yosys~\cite{Yosys} (v0.9) as the front-end engine. Yosys collaborates with ABC~\cite{abc} for logic synthesis, producing a minimized netlist for  PPA criteria. Graph processing uses networkx v2.6, and for graph-based ML, we use pytorch v1.9 and pytorch-geometric v1.7.0.
Area and timing data post-technology mapping from AIGs are collected using NanGate 45nm technology library and ``5K heavy” wireload model. Our dataset generation has three stages: (1) RTL synthesis, (2) Graph-level processing, and (3) Preprocessing for ML.
\subsubsection{RTL Synthesis} Logic synthesis begins with IP specifications (Verilog/VHDL). Yosys optimizes the sequential IP part, while ABC handles logic optimization and technology mapping for the combinational section. ABC forms AIGs through structural hashing of the combinational design. User-defined synthesis recipes optimize independently. We provide automated scripts, preserving intermediate AIGs. AIGs conform to the BENCH file format~\cite{brglez_combinational_1989}. Intermediate and final AIGs differ due to synthesis transformations. We curated $K = 1500$ synthesis recipes, each containing $L = 20$ transformations.

\subsubsection{Graph level processing}
\label{subsubsec:graphLevelProcessing} 
A gate-level netlist parser translates BENCH files (IP's AIG representation) into equivalent GRAPHML files, preserving AIG traits. Nodes are 2-input AND gates, edges consist of inverters/buffers. Features include node type, inverted edge count, and edge type. We store KxL = 30,000 graph structures per IP.

\subsubsection{Pre-processing for ML-ready dataset} We prepare circuit data using pytorch-geometric APIs, creating data samples from AIG graphs, synthesis recipes, and area/delay information. The dataset is accessible through a customized dataloader, simplifying preprocessing, labeling, and transformation. A script for dataset partitioning is also provided. 

\subsection{OpenABC-D characterization}

\begin{figure}[!htb]
    \centering
    \hspace*{-0.30in}
    \subfloat[\label{fig:h3}Top $10\%$]{\includegraphics[width=0.8\columnwidth, valign=c]{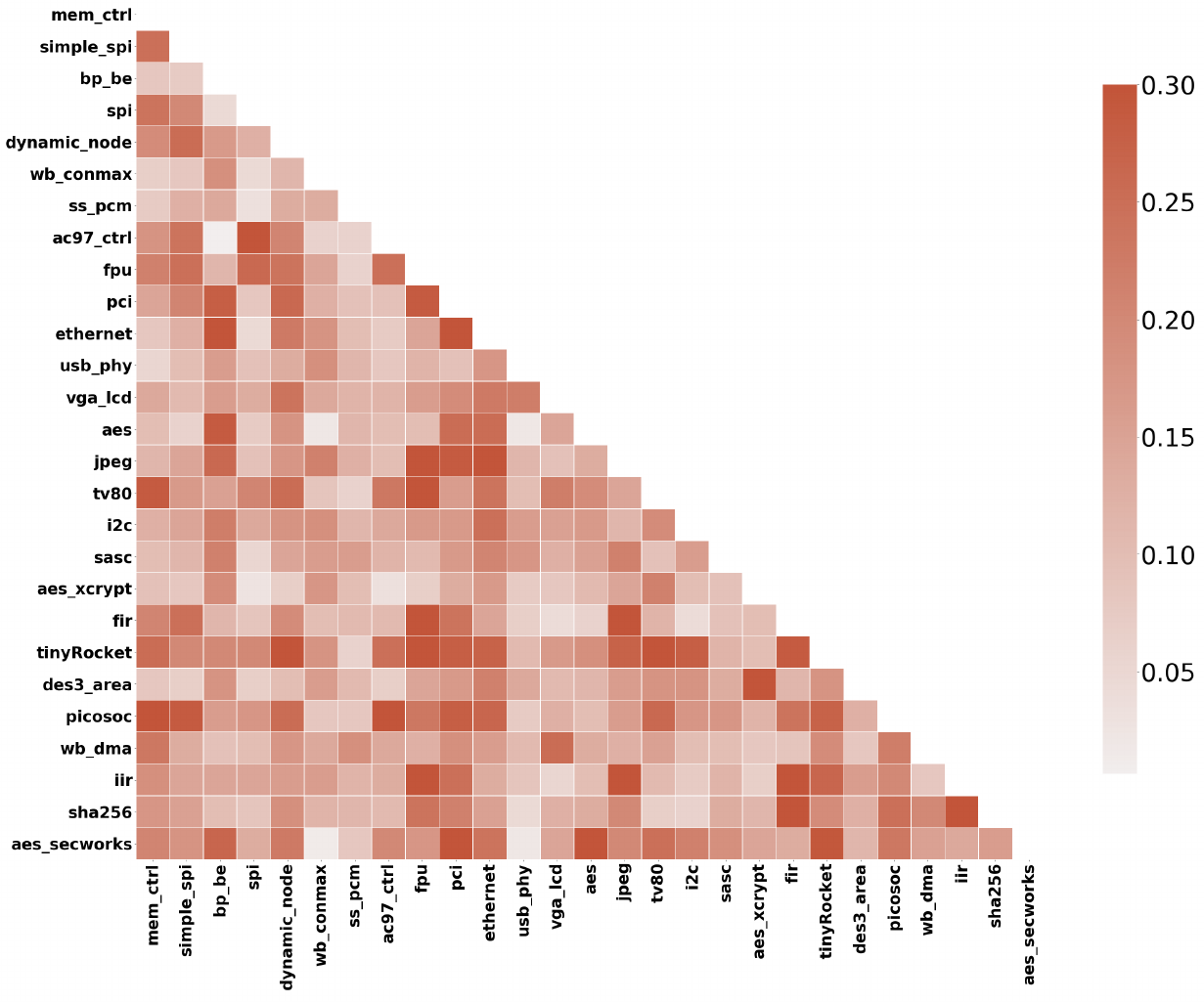}}
  \caption{Correlation plots of top $k\%$ synthesis recipes for IPs. Dark shade $\implies$ Similarity $\uparrow$ between top recipes for IP pairs.}
    \label{fig:synthesis-tops}
\end{figure}

For OpenABC-dataset, we use 29 open source IPs covering diverse functions. Lacking an all-encompassing dataset like ImageNet, we compiled IPs from MIT LL labs CEP~\cite{mitll-cep}, OpenCores~\cite{opencores}, and IWLS~\cite{albrecht2005iwls}. Unlike ISCAS~\cite{iscas85,iscas89} and EPFL~\cite{amaru2015epfl} benchmarks focused on limited function IPs, these benchmarks are more diverse. This diversity of IPs corresponds to a rich spectrum of AIG structures, mirroring real-world hardware designs. The benchmark characteristics are outlined in Table~\ref{tab:structuralcharacteristics}, spanning bus communication, processors, signal processing cores, accelerators, and controllers.
We used 1500 synthesis recipes, each with 20 steps, saving 20 AIGs per recipe for each IP. These recipes were generated by uniform sampling, necessitating over 200,000 computation hours on Intel processors for labeling. Analyzing the top 150 recipes (10\%), their similarity remains under 30\% (Fig.~\ref{fig:synthesis-tops}), highlighting diverse graph structures and transformation sequences.


\noindent{\textbf{AIG encoding}}: Following  Section~\ref{subsubsec:graphLevelProcessing}, we transform BENCH into GRAPHML using an adjacency matrix to retain node and edge attributes. Node features are numerically encoded: (1) node type and (2) count of preceding inverter edges. Edge encoding uses 0 for original and 1 for inverted signals.
\noindent{\textbf{Synthesis recipe encoding}}: Each synthesis recipe is uniquely encoded by a 20-dimensional synthesis ID vector.
\noindent{\textbf{Data sample creation}}: Data samples merge AIG and synthesis flow encodings with labels. Sample names like ``aes\_syn149\_step15.pt" include IP, recipe ID, and post-recipe AIG state after 15 transformations. Labels encompass primary input/output counts, AIG details (\# of nodes, inverted edges, depth), IP function, post-recipe AIG node count, and post-technology mapping area and delay. 

\begin{table}[!tb]
\centering
\scriptsize
\setlength\tabcolsep{1pt}
\resizebox{\columnwidth}{!}{%
\begin{tabular}{@{}lllllllc@{}}
\toprule
 & \multicolumn{7}{c}{Characteristics of Benchmarks} \\ 
 \cmidrule(l){2-8} 
 \multirow{-2}{*}{IP} & PI & PO & N & E & I & D & Function \\ \midrule
\rt{spi~\cite{opencores}}  & \rt{254} & \rt{238} & \rt{4219} & \rt{8676} & \rt{5524} & \rt{35} &  \rt{Serial peripheral interface}\\
\rt{i2c\cite{opencores}}  & \rt{177} & \rt{128} & \rt{1169} & \rt{2466} & \rt{1188} & \rt{15} & \rt{Bidirectional serial bus protocol}\\
\rt{ss\_pcm\cite{opencores}} & \rt{104} & \rt{90} & \rt{462} & \rt{896} & \rt{434} & \rt{10} & \rt{Single slot PCM}\\
\rt{usb\_phy\cite{opencores}} & \rt{132} & \rt{90} & \rt{487} & \rt{1064} & \rt{513} & \rt{10} & \rt{USB PHY 1.1}\\
\rt{sasc\cite{opencores}} & \rt{135} & \rt{125} & \rt{613} & \rt{1351} & \rt{788} & \rt{9} &  \rt{Simple asynch serial controller}\\
\rt{wb\_dma\cite{opencores}} & \rt{828} & \rt{702} & \rt{4587} & \rt{9876} & \rt{4768} & \rt{29} & \rt{Wishbone DMA/Bridge} \\
\rt{simple\_spi\cite{opencores}} & \rt{164} & \rt{132} & \rt{930} & \rt{1992} & \rt{1084} & \rt{12} & \rt{MC68HC11E based SPI interface}\\
\rt{pci\cite{opencores}} &	\rt{3429} &	\rt{3157} &	\rt{19547} & \rt{42251}	& \rt{25719} &\rt{29} & \rt{PCI controller}\\
 \rt{wb\_conmax\cite{opencores}} &	\rt{2122} &	\rt{2075} &	\rt{47840} & \rt{97755}	& \rt{42138} &	\rt{24} & \rt{WISHBONE Conmax}  \\
\rt{ethernet\cite{opencores}} &	\rt{10731} & \rt{10422} & \rt{67164} & \rt{144750}	& \rt{86799} &	\rt{34} & \rt{Ethernet IP core}\\
 \midrule
 \gt{ac97\_ctrl\cite{opencores}} &	\gt{2339} &	\gt{2137} &	\gt{11464} & \gt{25065} & \gt{14326} &	\gt{11} &  \gt{Wishbone ac97} \\
 \gt{mem\_ctrl\cite{opencores}} &	\gt{1187} &	\gt{962}& \gt{16307} & \gt{37146} & \gt{18092} &	\gt{36} & \gt{Wishbone mem controller}\\
 \gt{bp\_be\cite{bpsoc}} &	\gt{11592} & \gt{8413} & \gt{82514} & \gt{173441}	& \gt{109608} &	\gt{86} & \gt{Black parrot RISCV processor engine}\\
 \gt{vga\_lcd\cite{opencores}} & \gt{17322} & \gt{17063} &	\gt{105334} & \gt{227731} & \gt{141037} & \gt{23} & \gt{Wishbone enhanced VGA/LCD controller} \\
\midrule
\bt{des3\_area\cite{opencores}} & \bt{303} & \bt{64} & \bt{4971} & \bt{10006} & \bt{4686} & \bt{30} & \bt{DES3 encrypt/decrypt}\\
\bt{aes\cite{opencores}} & \bt{683} & \bt{529} & \bt{28925} & \bt{58379} & \bt{20494} & \bt{27} &  \bt{AES (LUT-based)}\\
\bt{sha256\cite{mitll-cep}} &	\bt{1943} &	\bt{1042} &	\bt{15816} & \bt{32674}	& \bt{18459} &	\bt{76} & \bt{SHA256 hash} 	\\
\bt{aes\_xcrypt\cite{aesxcrypt}} &	\bt{1975} &	\bt{1805} &	\bt{45840} & \bt{93485}	& \bt{36180} &	\bt{43} & \bt{AES-128/192/256}\\	
\bt{aes\_secworks\cite{aessecworks}}  &	\bt{3087} &	\bt{2604} &	\bt{40778} & \bt{84160}	&\bt{45391} &	\bt{42} & \bt{AES-128  (simple)} \\	
\midrule
\mat{fir\cite{mitll-cep}} & \mat{410} & \mat{351} & \mat{4558} & \mat{9467} & \mat{5696} & \mat{47} & \mat{FIR filter}\\
\mat{iir\cite{mitll-cep}} & \mat{494} & \mat{441} & \mat{6978} & \mat{14397} & \mat{8596} & \mat{73} & \mat{IIR filter} \\
\mat{jpeg\cite{opencores}} & \mat{4962} & \mat{4789} &	\mat{114771} &	\mat{234331} & \mat{146080} &	\mat{40} & \mat{JPEG encoder}\\
\mat{idft\cite{mitll-cep}} & \mat{37603} &	\mat{37419} & \mat{241552} & \mat{520523} & \mat{317210} &	\mat{43} & \mat{Inverse DFT}  \\
\mat{dft\cite{mitll-cep}} &	\mat{37597} & \mat{37417} &	\mat{245046} &	\mat{527509} & \mat{322206} &	\mat{43} & \mat{DFT design}\\

\midrule
\vt{tv80\cite{opencores}} &	\vt{636}	& \vt{361} & \vt{11328} & \vt{23017}	& \vt{11653} &	\vt{54} & \vt{TV80 8-Bit Microprocessor}  \\
\vt{tiny\_rocket\cite{ajayi2019openroad}} &	\vt{4561} &	\vt{4181} &	\vt{52315} & \vt{108811}	& \vt{67410} & \vt{80} & \vt{32-bit tiny riscv core}\\
\vt{fpu\cite{balkind2016openpiton}} &	\vt{632} &	\vt{409} &	\vt{29623} & \vt{59655}	& \vt{37142} &	\vt{819} & \vt{OpenSparc T1 floating point unit} \\
\bst{picosoc\cite{balkind2016openpiton}} &	\bst{11302} &	\bst{10797} &	\bst{82945} & \bst{176687}	& \bst{107637} &	\bst{43} & \bst{SoC with PicoRV32 riscv} \\
\fgt{dynamic\_node\cite{ajayi2019openroad}}  & \fgt{2708} & \fgt{2575} & \fgt{18094} & \fgt{38763} & \fgt{23377} & \fgt{33}  & \fgt{OpenPiton NoC architecture}\\


\bottomrule
\end{tabular}
}
\caption{Open source IP characteristics (unoptimized).  Primary Inputs (PI), Primary outputs (PO), `Nodes (N), Edges (E), `Inverted edges (I), Netlist Depth (D). Color code: \rt{Communication/Bus protocol}, \gt{Controller}, \bt{Crypto}, \mat{DSP}, \vt{Processor}, \bst{Processor$+$control}, \fgt{Control$+$Communication}\label{tab:structuralcharacteristics}}
\end{table}

\subsection{OpenABC-D Dataset Benchmarking}
The OpenABC-D dataset functions as an ML-driven EDA task benchmark in logic synthesis. It supports supervised learning tasks such as predicting synthesis outcome quality (For example, \% of nodes optimized, longest path) as shown in our findings (Fig.~\ref{fig:synthesis-tops}). Diverse IPs have varied effects of synthesis recipes, indicating the lack of a universal recipe. Predicting PPA for a recipe and IP can expedite recipe improvements, using metrics like normalized AIG node count for post-synthesis area.

\noindent{\textbf{1) Predict Unseen Recipe PPA (Transductive)}}: Can we predict synthesis outcome quality for a specific IP and recipe? Our model is trained on IP netlists and AIG outputs from 1000 recipes (29 × 1000 = 29000 samples). Evaluation checks if the model predicts AIG node count post-synthesis for 500 untested recipes. This reflects real scenarios where expert recipes are tested on IP blocks for quick PPA prediction for new recipes.


\noindent{\textbf{2) Predict Unseen IP PPA (Inductive)}}: Our model trains on PPA data from smaller IPs and predicts PPA for larger unseen IPs using an IP and recipe. This mirrors reality where synthesizing large IPs takes weeks and smaller ones are faster. We check if models trained on smaller IPs can predict outcomes for larger ones. The training set has 16 smaller IPs, and inference has 8 large IPs.

\begin{table}[t]
\centering
\caption{GCN architecture. I: Input dimesion, L1, L2: GCN dimension: f: \# filters, k: kernels, s: stride, \# l: \#FC layers\label{tab:network-architecture-params}}
\resizebox{\columnwidth}{!}{%
\begin{tabular}{@{}ccccccccccc@{}}
\toprule
\multirow{2}{*}{Net} & \multicolumn{4}{c}{AIG Embedding} & \multicolumn{4}{c}{Recipe Encoding} & \multicolumn{2}{c}{FC Layers} \\ \cmidrule(lr){2-5}\cmidrule(lr){6-9}\cmidrule(lr){10-11} 
 & I & L1 & L2 & Pool & I & \#f & k & s & \# l & arch\\ \midrule
Net1 & 4 & 128 & 128 & Max+Mean & 60 & 3 & 6,9,12 & 3 & 3 & 310-128-128-1\\
Net2 & 4 & 64 & 64 & Max+Mean & 60 & 4 & 12,15,18,21 & 3 & 4 & 190-512-512-512-1\\
Net3 & 4 & 64 & 64 & Max+Mean & 60 & 4 & 21,24,27,30 & 3 & 4 & 190-512-512-512-1\\
\bottomrule
\end{tabular}%
}
\end{table}

\subsection{Model architecture and hyperparameters}

\begin{figure}[t!]
    \centering
     \subfloat[][Network architecture]{\includegraphics[width=0.5\columnwidth]{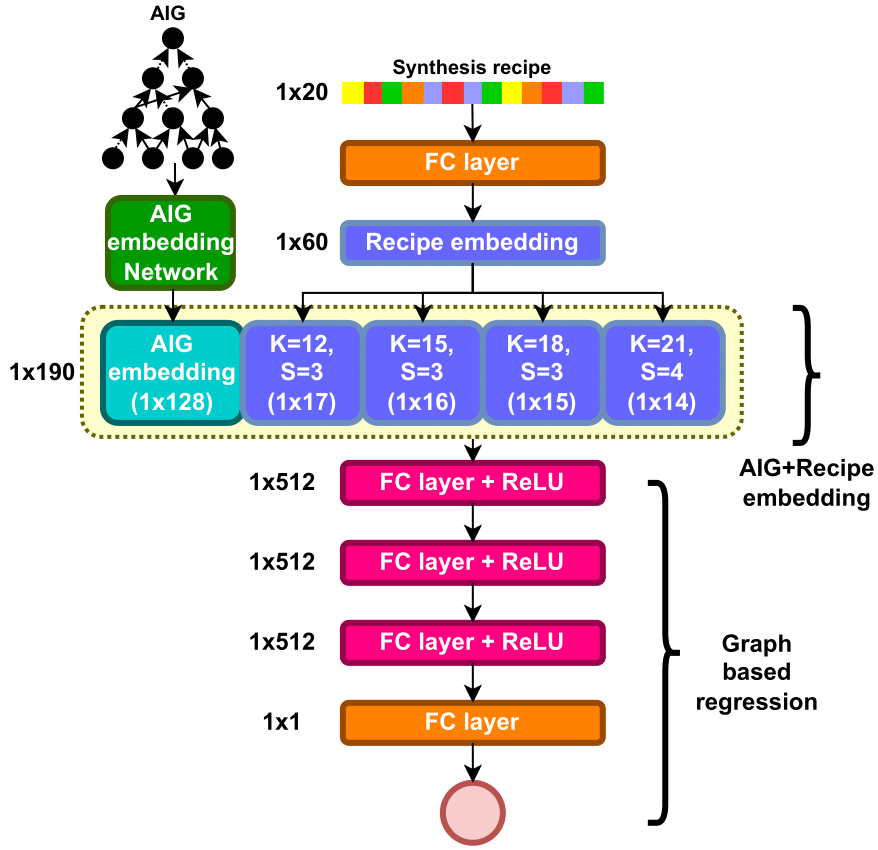}}
     \subfloat[][AIG embedding Network]{\includegraphics[width=0.5\columnwidth]{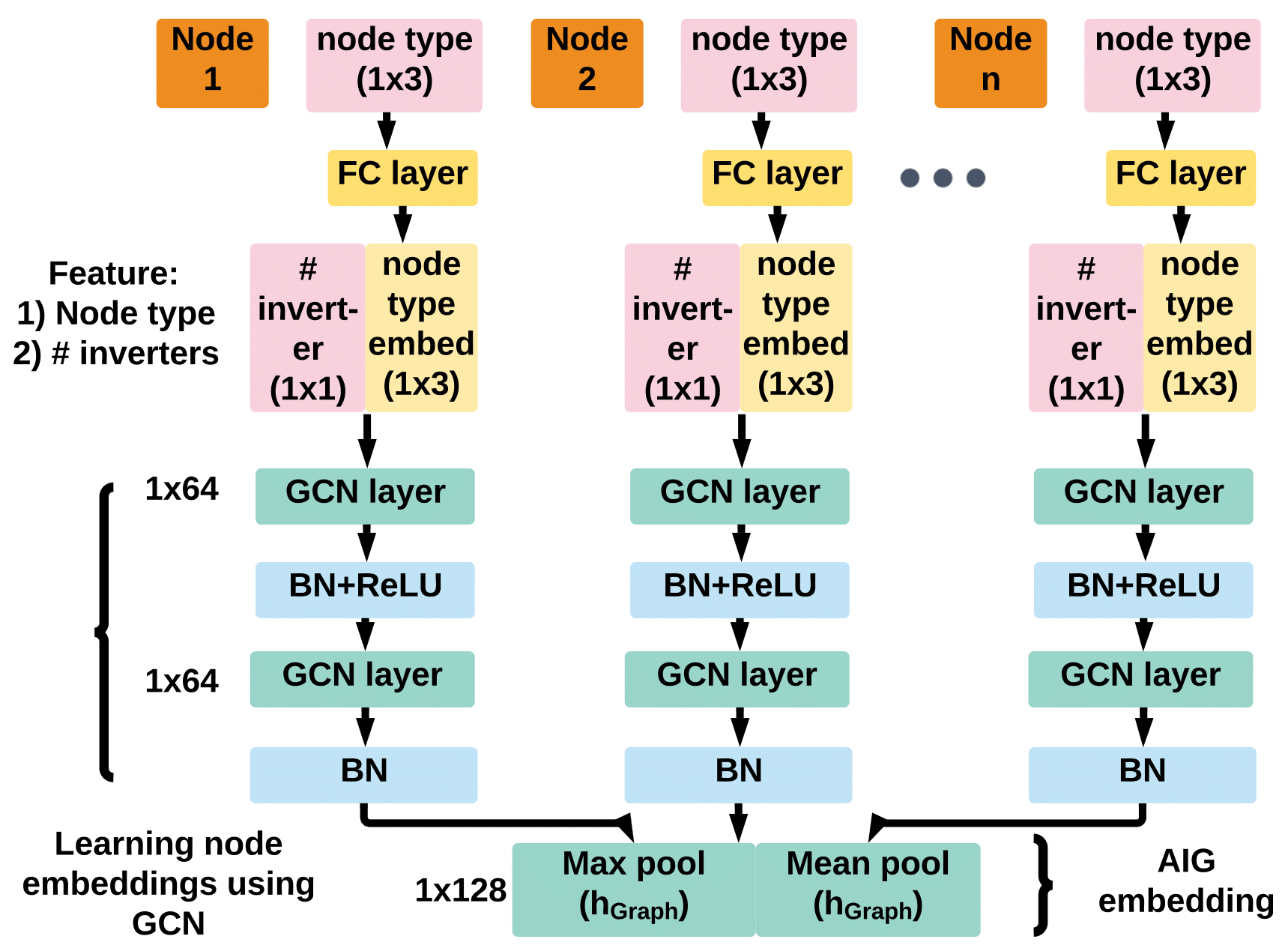}} \\
    \caption{Architecture for {PPA} prediction. $GCN$: Graph convolution. $FC$: fully connected. $BN$: Batch normalization.}
    \label{fig:OpenABCDnetwork}
\end{figure}

For all tasks, our model conducts graph-level predictions via synthesis recipe encoding. To evaluate graph convolution networks (GCN), we adopted a simple architecture (Fig.~\ref{fig:OpenABCDnetwork}a). This setup features a two-layer GCN with the AIG as input. The GCN captures node-level embeddings, while a graph-level embedding is obtained through global max pooling or average pooling. Synthesis recipe encoding involves a linear layer followed by 1D convolution. The convolution layer's kernel size and stride serve as adjustable hyperparameters. Graph-level and synthesis recipe embeddings are concatenated and fed through fully connected layers for regression. Table~\ref{tab:network-architecture-params} outlines three model configurations (hyperparameter setups). Across setups, we used batch size=64, initial learning rate=0.001, and conducted 80 training epochs, employing the Adam optimizer.

\subsection{Experimental results and analysis} We consider mean squared error (MSE) metric to evaluate model effectiveness. We presented our results on baseline networks in Table 6 for all the downstream tasks.

\begin{figure}[h]
    \centering
    \subfloat[\label{fig:aes_secworks_1}][]{\includegraphics[width=0.33\columnwidth, valign=c]{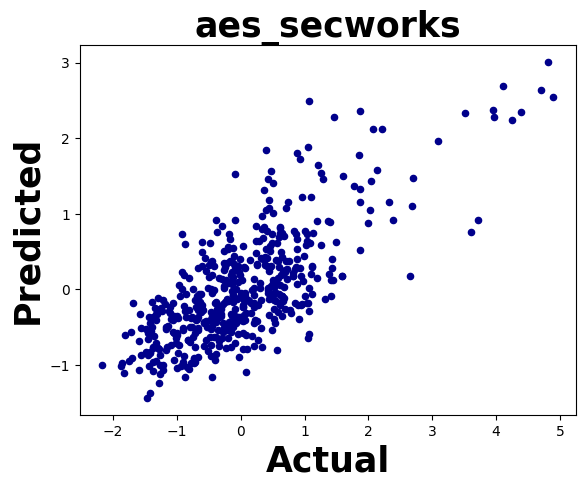}}
    \subfloat[\label{fig:tiny_rocket_1}]{\includegraphics[width=0.33\columnwidth, valign=c]{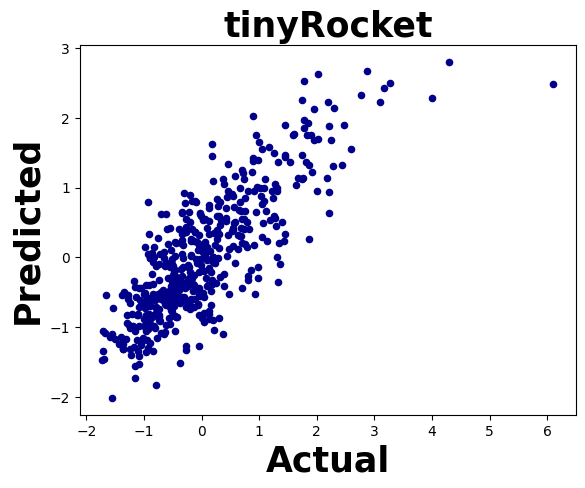}}
    \subfloat[\label{fig:tv_80_1}]{\includegraphics[width=0.33\columnwidth, valign=c]{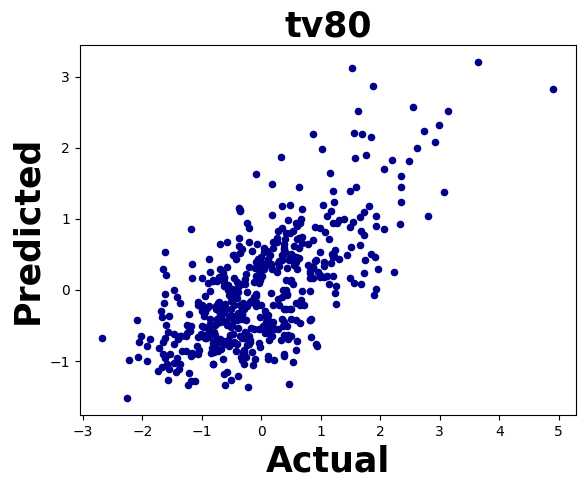}}\\ \vspace*{-0.1in}
    \subfloat[]{\includegraphics[width=0.33\columnwidth, valign=c]{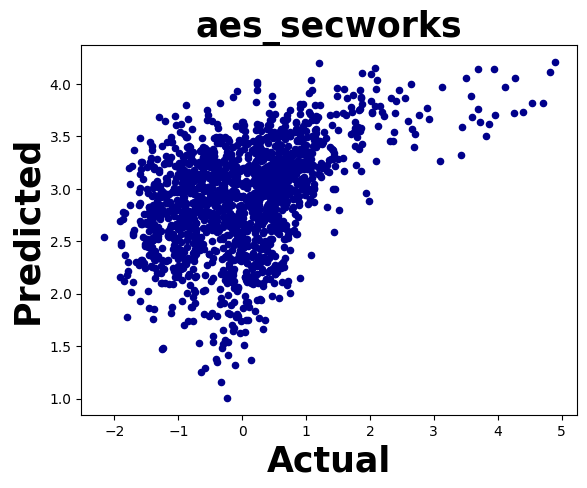}}
    \subfloat[]{\includegraphics[width=0.33\columnwidth, valign=c]{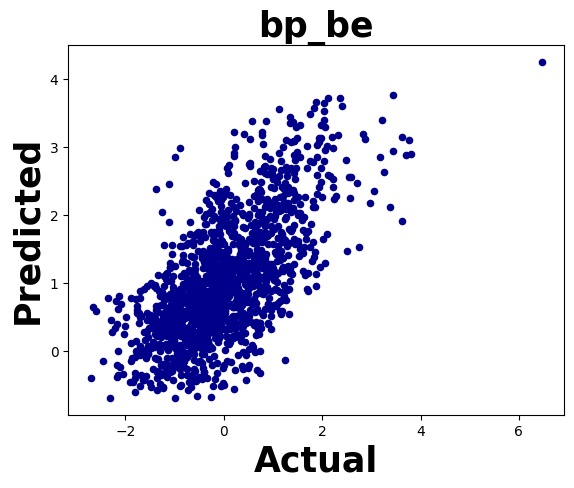}}
    \subfloat[]{\includegraphics[width=0.33\columnwidth, valign=c]{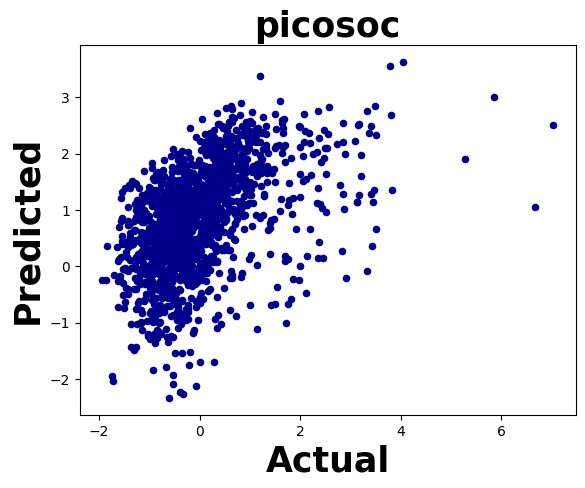}} 
    \vspace*{-0.1in}
   \caption{Net 1 for PPA Task 1 (top row) and 2 (bottom row)}
    \label{fig:net1_set1}
\end{figure}

\begin{table}[]
\centering
\resizebox{0.6\columnwidth}{!}{%
\begin{tabular}{@{}cccc@{}}
\toprule
\multirow{2}{*}{Task} & \multicolumn{3}{c}{Test MSE on baseline networks}\\
\cmidrule(lr){2-4}
& Net1 & Net2 & Net3 \\ \midrule
Task-1 & $0.648\pm0.05$  & $0.815\pm0.02$  & $\mathbf{0.579\pm0.02}$ \\

Task-2 & $10.59\pm2.78$ & $1.236\pm0.15$ & $\mathbf{1.47\pm0.14}$ \\

\bottomrule
\end{tabular}
}
\caption{Benchmarking \ac{GCN} models for PPA prediction tasks}
\label{tab:qorsynthesisrecipe}
\end{table}

\subsubsection{Predicting PPA for Unseen Synthesis Recipes} We present scatter plots illustrating predicted versus actual normalized node counts in optimized netlists (Fig.~\ref{fig:net1_set1}(top row). The plots closely align with the y = x trend, indicating strong performance in predicting PPA for unknown recipes. Kernel filters learn and encapsulate features within synthesis sub sequence, contributing to their effective performance. A slight deviation in scatter plots for IPs (fir, iir, and mem\_ctrl) is worth noting. Net3 shows improved accuracy on these IPs compared to net1 and net2. Net1 outperforms net2 on mem\_ctrl.


\subsubsection{PPA Prediction on Unseen IP} In the train-test split strategy where IPs are unknown during testing, different networks exhibit variable performance for various test IPs (refer to Fig.~\ref{fig:net1_set1}(bottom row)). IPs like aes\_xcrypt and wb\_conmax consistently yield subpar results, indicating disparities in AIG embeddings. Conversely, results for IPs like bp\_be, tinyRocket, and picosoc align closely with training data, highlighting the network's capacity to draw insights from graph structures.


\section{Conclusion}

VeriGen and OpenABC-D are datasets tailored for advancing ML in electronic design automation (EDA). VeriGen is a comprehensive corpus of real-world Verilog code useful to train models that can produce high-quality Verilog. We developed an open evaluation pipeline for assessing syntactic and functional integrity.  OpenABC-D provides 870,000 synthesized netlists with area and delay details. OpenABC-D and its open evaluation pipeline, sets the stage for ML models for tasks like design space exploration and logic synthesis recipe recommendation. The two datasets address the need for standardized data sets in EDA and are open-sourced to the community to benefit research. 

\section*{Acknowledgment}
This work was partly funded by Qualcomm, Intel, and Synopsys. The views expressed are solely those of the author(s) and do not represent the sponsors.



\bibliographystyle{IEEEtran}
\bibliography{main}


\end{document}